\documentclass{article}

\usepackage[final, nonatbib]{neurips_2022}

\usepackage[utf8]{inputenc} 
\usepackage[T1]{fontenc}    
\usepackage{hyperref}       
\usepackage{url}            
\usepackage{booktabs}       
\usepackage{amsfonts}       
\usepackage{nicefrac}       
\usepackage{microtype}      
\usepackage{xcolor}         
\usepackage{graphicx}

\usepackage{enumitem}
\usepackage{amsmath}
\usepackage{amssymb}

\usepackage{caption,subcaption}

\usepackage{algorithmic}
\usepackage{algorithm}          

\DeclareMathOperator*{\argmin}{arg\,min}

\newcommand\grid[1]{\includegraphics{figures/imagenet/#1}}

\title{Learning Probabilistic Models from Generator Latent Spaces with Hat EBM}

\author{%
  Mitch Hill \\
  OPPO US Research Center\\
  \texttt{mitch.hill@innopeaktech.com} \\
  \And
  Erik Nijkamp \\
  Salesforce Research \\
  \texttt{erik.nijkamp@salesforce.com} \\
  \AND
  Jonathan Mitchell \\
  University of California, Los Angeles \\
  \texttt{jcmitchell@ucla.edu} \\
  \And
  Bo Pang \\
  Salesforce Research \\
  \texttt{b.pang@salesforce.com} \\
  \And
  Song-Chun Zhu \\
  University of California, Los Angeles\\
  \texttt{sczhu@stat.ucla.edu} \\
}

\begin{document}

\maketitle

\begin{abstract}
This work proposes a method for using any generator network as the foundation of an Energy-Based Model (EBM). Our formulation posits that observed images are the sum of unobserved latent variables passed through the generator network and a residual random variable that spans the gap between the generator output and the image manifold. One can then define an EBM that includes the generator as part of its forward pass, which we call the Hat EBM. The model can be trained without inferring the latent variables of the observed data or calculating the generator Jacobian determinant. This enables explicit probabilistic modeling of the output distribution of any type of generator network. Experiments show strong performance of the proposed method on (1) unconditional ImageNet synthesis at 128$\times$128 resolution, (2) refining the output of existing generators, and (3) learning EBMs that incorporate non-probabilistic generators. Code and pretrained models to reproduce our results are available at \url{https://github.com/point0bar1/hat-ebm}.
\end{abstract}

\section{Introduction}

Generator networks~\cite{kingma2013auto,goodfellow2014generative} which transform a latent distribution to a complex observed distribution (e.g., images, videos) are found across different deep generative models. One limitation of generator networks is the difficulty of obtaining an explicit representation of the probability distribution defined by the output of the network after transformation of the latent space. For generators from Generative Adversarial Networks (GANs) ~\cite{goodfellow2014generative,radford2015unsupervised} and Variational Autoencoders (VAEs) ~\cite{kingma2013auto,rezende2014stochastic} where the latent states corresponding to realistic images follow a trivial distribution (e.g., isotropic Gaussian), the difficulty of obtaining image space probabilities lies in calculating the log determinant of the generator Jacobian which is needed to perform density change-of-variables. For other generator models such as a deterministic autoencoder, there might not be a natural way to generate probabilistic samples from the latent space that correspond to realistic images. When the latent space has smaller dimension than the image space, there is further difficulty in describing the generator image distribution since the distribution measure must be confined to the manifold of the generator output.

\begin{figure*}
    \centering
    \captionsetup[subfigure]{labelformat=empty}
    \subfloat[]{%
        \includegraphics[width=.425\textwidth]{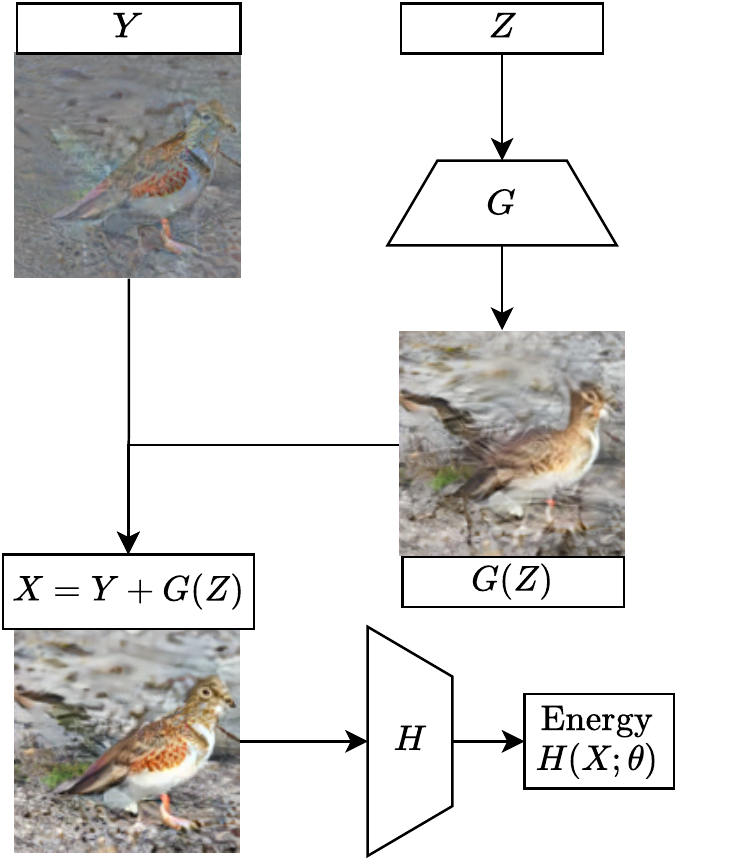} \quad
    }
    \hspace{0.01cm}
    \subfloat[]{%
        \resizebox{.5\textwidth}{!}{%
            \setlength{\tabcolsep}{0.25em}
            \def\arraystretch{1.5}
            \begin{tabular}{cccc}
                \grid{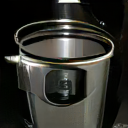} & \grid{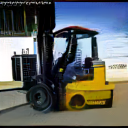} & \grid{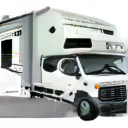}& \grid{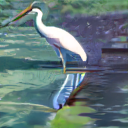} \\
                \grid{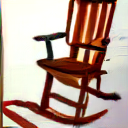} & \grid{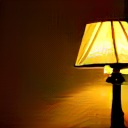} & \grid{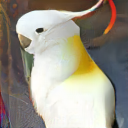}& \grid{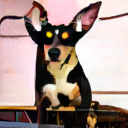} \\
                \grid{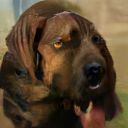} & \grid{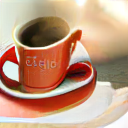} & \grid{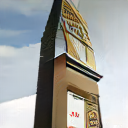}& \grid{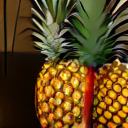} \\
                \grid{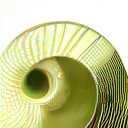} & \grid{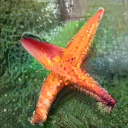} & \grid{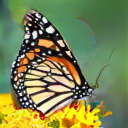}& \grid{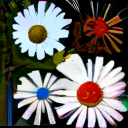}
            \end{tabular}
        }
    }
    \vspace*{-5mm}
    \caption{\emph{Left:} The Hat EBM model takes a joint input $(Y, Z)$ where $Y$ is a residual image and $Z$ is a latent vector. An image is generated using $X = Y+G(Z)$ for a generator $G(Z)$, and the image is passed to the hat network $H(X;\theta)$ to obtain the energy of the pair $(Y, Z)$. This allows for principled probabilistic learning which can incorporate the latent space of any generator. \emph{Right:} Unconditional ImageNet 128$\times$128 samples generated by a Hat EBM.}
    \vspace*{-2mm}
    \label{fig:main_fig}
\end{figure*}

This work proposes a method for using a generator network as the foundation for an Energy-Based Model (EBM). The generator network is concatenated with a \emph{hat network} that takes an image input and outputs a scalar. Before the generator output is fed into the hat network, the generated image is adjusted by adding a residual image that spans the gap between the generator output and the image manifold. The total function, including the generator, addition of the residual image, and the hat network, is called the \emph{Hat EBM}. This formulation allows us to define an EBM which can incorporate the generator latent space as part of its image space density and MCMC sampling process. Figure~\ref{fig:main_fig} displays a diagram of the Hat EBM and selected Hat EBM samples for unconditional ImageNet at resolution $128\times128$. Our method is general and applies to any generator model, in contrast to existing methods for converting generator networks to EBMs which only apply to a specific generator model like a GAN~\cite{che2020your} or VAE~\cite{xiao2021vaebm}. Additionally, the learning method does not require inference of latent states of the observed data as used in related works \cite{kingma2013auto,xie_coop,han2019divergence,pang2020learning}.

The Hat EBM formulation enables applications with frozen pretrained generator networks such as learning an EBM to refine samples from generators with probabilistic latent spaces (e.g. GANs) and synthesizing samples using non-probabilistic generators (e.g. deterministic autoencoders). We also propose a self-contained learning method that extends cooperative learning of EBM and generator networks \cite{xie_coop} to achieve high-quality synthesis. Our main contributions are summarized below.

\begin{itemize}[leftmargin=*]
    \item We introduce a method for defining a Hat EBM that incorporates a generator network as part of its forward pass. This EBM includes the generator latent space as part of MCMC sampling.
    \item We show that our method can refine samples from pretrained GAN generators and sample from the latent space of deterministic autoencoders which are originally incompatible with sampling.
    \item We propose a self-contained Hat EBM learning method that trains both a generator and energy network from scratch. This enables us to achieve an FID score of 29.2 on unconditional ImageNet at resolution 128$\times$128, demonstrating the EBMs are competitive with state-of-the-art generative models on complex and high resolution datasets.
\end{itemize}

\section{Related Work}

This section highlights important related methods for EBM learning. A more thorough discussion of related work can be found in Appendix~\ref{app:ebm_discussion}.

\noindent \textbf{EBM.} An EBM defines an unnormalized density or equivalently a Gibbs-Boltzmann distribution. Prototypes include exponential family distributions, Boltzmann machines \cite{hinton1985boltzmann, salakhutdinov2009deep}, and the FRAME (Filters, Random field, And Maximum Entropy) model \cite{zhu1998frame}. Recent work has introduced the EBM with a ConvNet potential~\cite{xie2017synthesizing, xie2018learning}. This dramatically increases the model capacity and enables strong image synthesis performance~\cite{nijkamp2019learning, du2019implicit} and adversarial robustness~\cite{hill2021stochastic}. Several works investigate training an EBM in tandem with an auxiliary model. Kim and Bengio \cite{kim2016deep} jointly train an EBM and generator without MCMC by using samples from the generator as direct approximations of the EBM density. The EGAN \cite{dai2017calibrating} builds on this method by introducing a maximum entropy objective to improve generator training. A similar approach is explored by the VERA model \cite{nomcmcforme}. Cooperative learning \cite{xie2021cooperative} trains the EBM and generator by using the generator to initialize samples needed to train the EBM and uses reconstruction loss between generator and EBM samples to learn the generator. The Flow Contrastive EBM \cite{gao2020flow} learns an EBM using Noise Contrastive Estimation with an auxiliary flow model. 
 
 
 \noindent \textbf{Latent Space EBM.} EBMs in the data space can be highly multi-modal, and MCMC sampling can be difficult \cite{xie2016theory, nijkamp2019learning, du2019implicit}. Recent works \cite{pang2020learning, pang2021latent} explore learning an EBM in latent space, which is then mapped to the data space with a learned generator. The energy landscape in the latent space is smoother and less multi-modal because it occupies much lower dimensional space and stands upon an expressive generator. These works define a prior EBM in the latent space as a correction of the non-informative uniform prior or isotropic Gaussian prior. To learn the model, one needs to infer the posterior of the latent variables. Posterior inference given such a complicated model is non-trivial. One needs to either design a sophisticated amortized inference network or run expensive MCMC. Our model also defines an EBM in the latent space, while its learning does not need posterior inference, making the learning much simpler and more scalable. Several works \cite{tanaka2019discriminator, che2020your, ansari2021refining} leverage a pretrained GAN to define an EBM in the latent space of the generator with a correction based on the discriminator, and shows improved synthesis quality over the pretrained GAN. The VAEBM \cite{xiao2021vaebm} uses a pretrained VAE to facilitate EBM learning. Our model can likewise be utilized to improve the quality of images from pretrained GAN or VAE generators. Our method is however more general since it can be used to obtain realistic samples from any pretrained generator, including non-probabilistic generators from deterministic autoencoders.     

\section{Formulation of Hat EBM}

This section presents the formulation of the Hat EBM energy function and the proposed learning procedure. We first review the fundamental equations of EBM learning. Then we introduce two variants of the Hat EBM: one that learns a joint distribution over the latent space and residual image space, and one that learns a distribution of residual images conditional on a fixed latent state from a known latent distribution. Finally, we propose a method for learning the hat network and generator network of a Hat EBM simultaneously so that our model can be used for self-contained image generation without the need for a pretrained generator.

\subsection{Review of EBM Learning} \label{sec:ebm_review}

We briefly review the main components of EBM learning following the standard method derived from works such as \cite{hinton2002poe, zhu1998frame, xie2016theory}. A deep EBM has the form
\begin{align}
p(x; \theta)=\frac{1}{\mathcal{Z}(\theta)} \exp\{-U(x;\theta)\} \label{eqn:ebm}
\end{align}
where $U(x;\theta)$ is a deep neural network with weights $\theta$ and $\mathcal{Z}(\theta)$ is the intractable normalizing constant. Given a true but unknown data density $q(x)$, Maximum Likelihood learning uses the objective $\argmin_\theta D_{KL} (q(x) \, \| \, p(x;\theta))$, which can be minimized using the stochastic gradient
\begin{align}
\nabla \mathcal{L}(\theta) \approx \frac{1}{n}\sum_{i=1}^n \nabla_\theta U(X_i^+;\theta)-\frac{1}{n}\sum_{i=1}^{n} \nabla_\theta U(X_i^-;\theta)
\label{eqn:ebmlearning}
\end{align}
where the positive samples $\{X_i^+\}_{i=1}^n$ are a set of data samples and the negative samples $\{X_i^-\}_{i=1}^n$ are samples from the current model $p(x;\theta)$. To obtain the negative samples for a deep EBM, it is common to use MCMC sampling with the $K$ steps of Langevin equation
\begin{align}
X^{(k+1)} = X^{(k)} - \frac{\varepsilon^2}{2} \nabla_{X^{(k)}} U(X^{(k)} ;\theta) + \varepsilon V_k , \label{eqn:langevin}
\end{align}
where $\varepsilon$ is the step size and $V_k \sim N(0, I)$. The Langevin trajectories are initialized from a set of states $\{X^-_{i, 0}\}_{i=1}^n$ obtained from a certain initialization strategy.

\subsection{Hat EBM: Joint Distribution of Latent and Residual Image}
\label{sec:hat_joint}

This section presents our method for adapting a fixed generator network $G(z)$ to be part of an EBM, which we call the \emph{Hat EBM}. The Hat EBM defines the joint distribution of the random variable $Z \in \mathbb{R}^{m}$ in the $m$-dimensional latent space of the generator network and a random variable $Y \in \mathbb{R}^d$ in the $d$-dimensional image space. The joint energy has the form
\begin{equation}
    U(Y, Z ; \theta) = H (G(Z) + Y ; \theta)
    \label{eqn:hat-ebm}
\end{equation}
where $H(x ; \theta)$ is a neural network that takes an image $x\in \mathbb{R}^d$ as input and returns a scalar output. The weights of $H$ are given by $\theta$. We call $H$ the \emph{hat network} because it sits atop the generator $G$ to incorporate the generator latent space directly into the probabilistic model. 

The random variable $Y$ is meant to accommodate the gap between the output of $G(Z)$ and the image manifold, since in general we expect that $G(Z)$ contains an approximate but imperfect representation of the image distribution which can be refined by the residual state $Y$. In practice, we find that the majority of the appearance of a sampled image $X = G(Z) + Y$ comes from the generator output $G(Z)$ and not the residual image $Y$, indicating that the majority of the sampling behaviors of our model are determined by the latent space of $G$ (see Figure~\ref{fig:viz_fig}). 

An appealing aspect of the Hat EBM formulation is that we can learn the model without either calculating the log determinant of the Jacobian of $G(Z)$ as would be required for an energy of the form $U(G(Z); \theta)$, or inferring the $Z$ latent vectors associated with observed images $X$ as done in existing work on learning latent prior EBMs \cite{pang2020learning}. This is possible because we define the distribution of observed images $X$ by $X = G(Z) + Y$ where the pair $(Y, Z)$ is drawn from a true but unknown density $q(y, z)$. We can then use the Maximum Likelihood framework in Section~\ref{sec:ebm_review} to learn the weights $\theta$ of the hat network $H(x; \theta)$ by minimizing the objective $\argmin_\theta D_{KL} (q(y, z) || p(y, z;\theta))$ where
\begin{align}
    p(y, z; \theta) = \frac{1}{\mathcal{Z}(\theta)} \exp\{-H(G(z) + y;\theta)\}. \label{eqn:hat-ebm-prob}
\end{align}
One can obtain negative samples using alternating Langevin updates
\begin{align}
Y^{(k+1)} &= Y^{(k)} - \frac{\varepsilon_1^2}{2} \nabla_{Y^{(k)}} H(G(Z^{(k)}) + Y^{(k)} ; \theta) + \varepsilon_1 V_{k, 1} \label{eqn:langevin-hat-ebm-1} \\
Z^{(k+1)} &= Z^{(k)} - \frac{\varepsilon_2^2}{2} \nabla_{Z^{(k)}} H(G(Z^{(k)}) + Y^{(k+1)} ; \theta) + \varepsilon_2 V_{k, 2} \label{eqn:langevin-hat-ebm-2}
\end{align}
which switch off between updates with respect to $z$ and updates with respect to $y$. This sampling algorithm is essentially Metropolis-within-Gibbs since the Langevin update can be written as a Metropolis-Hastings step with Gaussian proposal, in which case \eqref{eqn:langevin-hat-ebm-1} and \eqref{eqn:langevin-hat-ebm-2} define a valid sampler for $p(y, z; \theta)$. Finally, updating $\theta$ can be accomplished using
\begin{align}
\nabla \mathcal{L}(\theta) \approx \frac{1}{n}\sum_{i=1}^n \nabla_\theta H(X_i^+;\theta)-\frac{1}{n}\sum_{i=1}^{n} \nabla_\theta H(G(Z_i^-) + Y_i^-;\theta)
\label{eqn:hat-ebm-learning}
\end{align}
where $X_i^+$ are observed samples and the pairs $(Y_i^-, Z_i^-)$ are obtained via MCMC. In our formulation the observed data $X_i^+$ are sufficient statistics for $H(x; \theta)$ and there is no need to infer the $(Y_i^+, Z_i^+)$ pairs for the positive samples when learning the weights of the hat network.

\subsection{Conditional Hat EBM: Residual Image Conditional on Latent Sample} \label{sec:cond-hat-ebm}

Next we present a conditional variant of the Hat EBM. While the previous version of the Hat EBM is applicable to any generator network $G(z)$, including generators from deterministic autoencoders which cannot typically be sampled from, the conditional version of the Hat EBM is tailored towards generator networks which map a trivial latent distribution to complex signals like images. For these kinds of generators, one can use the known latent distribution as an ancestral distribution and learn a conditional distribution of the residual image given a latent sample. We emphasize that the conditional Hat EBM can be used to learn an unconditional distribution of observed images $X$ and that the term \emph{conditional} refers to the relationship between the latent variables $Y$ and $Z$. Our experiments focus on modeling only observed images $X$ without conditional information such as labels or captions.

Suppose we use a trivial marginal distribution $p_0 (z)$ for $Z$ and a generator $G$ trained to produce realistic images from this latent distribution. In our experiments, $p_0$ is always $N(0, I)$. We can now define a conditional Hat EBM density $p(y | z ;\theta) = \frac{1}{\mathcal{Z}_z(\theta)} \exp\{ - H (G(z) + y ; \theta) \}$ and a joint density
\begin{align}
    p(y, z; \theta) = \frac{1}{\mathcal{Z}_z(\theta)} \, p_0(z) \exp\{-H(G(z) + y;\theta)\}. \label{eqn:cond-hat-ebm-prob}
\end{align}
In this case, we posit that observed images $X$ are generated according to $X = G(Z) + Y$ for some distribution $q(y, z) = p_0 (z) q(y|z)$. Obtaining the negative samples $(Z_i^-, Y_i^-)$ is done by first drawing $Z_i^-$ from $p_0 (z)$ and then obtaining $Y_i^- | Z_i^-$ by using Langevin updates on the conditional probability $p(y | z ;\theta)$. Note that $p(y | z; \theta) = p(y | G(z); \theta)$ because of the form of \eqref{eqn:cond-hat-ebm-prob}. We update $\theta$ using the same equation \eqref{eqn:hat-ebm-learning} as the joint Hat EBM because $X_i^+$ is still a sufficient statistic for learning $H(x; \theta)$ and because we do not need to infer the $Z_i^+$ for $X_i^+$ since the prior $p_0 (z)$ does not contain any model parameters. In practice we initialize Langevin sampling from $Y_0 =0$ and $Z \sim N(0, I)$, and perform $K$ steps of \eqref{eqn:langevin-hat-ebm-1} to draw a residual sample $Y_K$ while keeping $Z$ fixed.

\subsection{Learning the Hat Network and Generator in Tandem for Conditional Hat EBM}
\label{sec:hat_tandem}

Both formulations of the Hat EBM above assume that a pretrained generator network $G(z)$ is available as the basis for learning the Hat EBM. In order to use the Hat EBM as a self-contained learning process for image generation, we now propose a method to learn the weights $\phi$ of a generator network $G(z; \phi)$ and the weights $\theta$ of a hat network $H(x; \theta)$ simultaneously. Our method is based on the cooperative learning~\cite{xie_coop} strategy. We first briefly review the cooperative learning formulation, and then present the learning formulation for the Hat EBM. A key difference between the derivations is that the original cooperative learning method requires MCMC inference of the latent variable $\hat{Z}$ associated with an MCMC sample $X$ to train $G(z; \phi)$, while in our formulation the latent variable $Z$ is explicitly defined and does not need to be inferred. This enables major computational savings because we can bypass the expensive MCMC inference of $\hat{Z}$.

In the original cooperative learning~\cite{xie_coop}, the generator output $G(z; \phi)$ is trained to match the appearance of a Langevin chain $X_K$ sampled from the potential $U(x; \theta)$ and initialized from the state $X_0 = G(Z_0 ; \phi)$ where $Z_0 \sim N(0, I)$. The model defines the conditional density of an images $X$ given latents $Z \sim N(0, I)$ as $X | Z \sim N(G(Z; \phi), \tau^2 I)$ for some sufficiently small $\tau > 0$. Given a sampled state $X_K$, updating $G(z;\phi)$ requires inferring $Z | X_K$ using the latent Langevin equation
\begin{equation}
    Z_{k+1} = Z_k - \frac{\varepsilon^2}{2} \left(Z_k + \frac{1}{2\tau^2} \nabla_{Z_k} \| G(Z_k ; \phi) - X_K \|_2^2 \right) + \varepsilon V_k
\end{equation}
before updating $\phi$ using the Maximum Likelihood stochastic gradient
\begin{equation}
    \nabla \mathcal{L}_G (\phi) \approx \frac{1}{n} \sum_{i=1}^n \frac{1}{2 \tau^2} \nabla_\phi \| G(Z_{K, i} ; \phi) - X_{K, i} \|_2^2 \label{eqn:coop_update}
\end{equation}
where the $i$ index denotes member $i$ of a batch with size $n$. We note that the code released with the cooperative learning method does not infer the latent variable of observed images and $Z_0 \sim N(0, I)$ is used in place of $Z_K$ in the objective \eqref{eqn:coop_update}. In accordance with this approach, we find that inferring $Z_K$ often hurts model performance and leads to additional complication. The difficulty of inferring $Z | X_K$ and the omission of this step in practice leave an unresolved gap in the cooperative learning formulation. The Hat EBM generator update allows us to bypass the Langevin update for $Z$ without theoretical complications.

\begin{figure*}
    \centering
    \includegraphics[width=.5\textwidth]{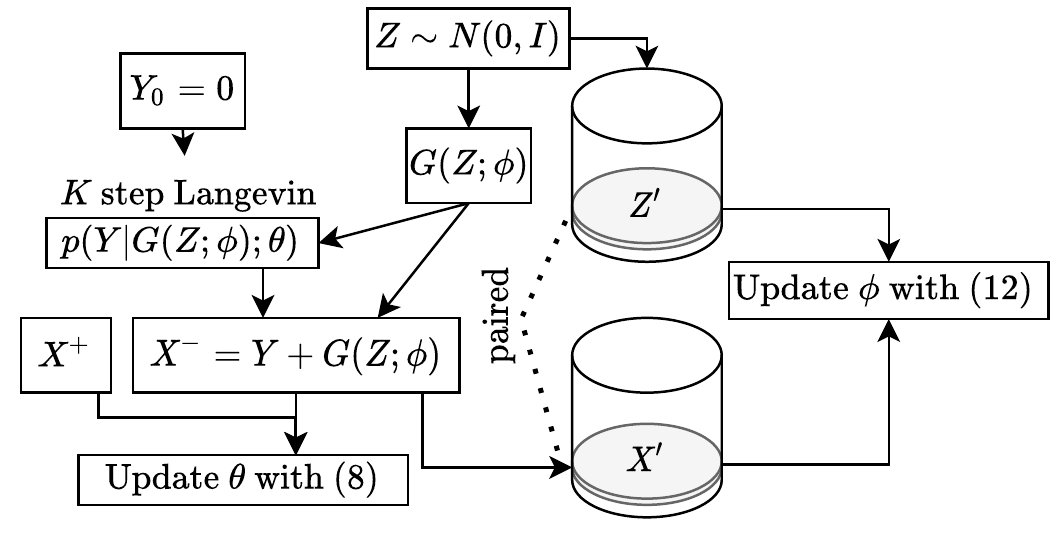}
    \caption{Visualization of tandem training method for hat network and generator. The left side illustrates training for the hat network $H(x;\theta)$. $Z$ is drawn from a latent distribution, $Y$ is initialized from the 0 image and updated according to $p(Y|Z;\theta, \phi)=p(Y|G(Z;\phi);\theta)$. Then data samples $X^+$ and negative samples $X^- = Y + G(Z;\phi)$ are used to update the weight $\theta$ of the hat network. On the right, pairs $(X', Z')$ from past hat network updates will be drawn randomly from a bank of states to update the weight $\phi$ of the generator. The bank memory $(X', Z')$ will then be overwritten by a new pair $(X, Z)$ from the current model.}
    \vspace*{-2mm}
    \label{fig:cond-hat-ebm}
\end{figure*}

To update the weights of a Hat EBM generator, we propose to train the generator to match the image samples produced by the current Hat EBM. Given the current generator weights $\phi_t$ and hat network weights $\theta_t$ at step $t$, we define our learnable model as $Z\sim N(0, I)$ and $X | Z \sim N(G(Z; \phi), \tau^2 I)$ and we define the true distribution of $(X, Z)$ as $Z\sim N(0, I)$ and $X = Y_K + G(Z; \phi_t)$ where $Y_K |Z$ is drawn from $K$ steps of Langevin sampling with Hat EBM density $p(y|z; \theta_t, \phi_t) \propto \exp\{-H(G(z;\phi_t) + y; \theta_t) \}$. Langevin updating is only used to obtain $Y_K$ while $Z$ remains fixed, as in the method from Section~\ref{sec:cond-hat-ebm}. Then $\phi$ can be updated using the Maximum Likelihood objective
\begin{equation}
    \phi_{t+1} = \argmin_\phi \frac{1}{2 \tau^2} E_{p_0 (z) p(y | z; \theta_t, \phi_t)} \left[ \| G(Z ; \phi) - (Y + G(Z ; \phi_t)) \|_2^2 \right] . \label{eqn:coop_update_hat}
\end{equation}
Conceptually, this loss function encourages $G(Z; \phi)$ to closely match the appearance of samples $X = Y + G(Z; \phi_t)$ created from a fixed generator $G(Z; \phi_t)$ and fixed hat network $H(x; \theta_t)$. In other words, the generator update should achieve $G(Z; \phi_{t+1}) \approx Y + G(Z; \phi_t)$ so that the updated generator absorbs the residual image $Y|Z$ from the current Hat EBM. Once the generator absorbs the current Hat EBM residual, the hat network update should learn to synthesize residual images that refine the the generator output to be more similar to the observed data. Like the original cooperative learning method, our generator is trained using only synthetic images and no observed data images are used to update $\phi$.

Ideal training of $H$ and $G$ would alternate between using the gradient of \eqref{eqn:coop_update_hat} until generator convergence is reached and using the gradient \eqref{eqn:hat-ebm-learning} to update the hat network. In practice we implement the minimization in \eqref{eqn:coop_update_hat} using only one gradient update initialized from $\phi = \phi_t$ to obtain $\phi_{t+1}$ rather than training until full convergence. This is done to increase training efficiency and to avoid maintaining a separate copy of generator weights for the fixed network $G(z; \phi_t)$. 

In our experiments we observe that using a single gradient update with the objective \eqref{eqn:coop_update_hat} has limited success because the generator output can become too closely tethered to biases of the current hat network. We find the same problems with the original cooperative learning objective \eqref{eqn:coop_update} (see Appendix~\ref{app:hist_gen}). To overcome these problems, we choose to train $G(z; \phi)$ at time $t+1$ to match the historical distribution of hat networks $H(x; \theta_{t_\ell})$ and generators $G(z; \phi_{t_\ell})$ for a selection of past epochs $t_1, t_2, \dots t_L \le t$ instead of training the generator to match the distribution of the current hat network $H(x; \theta_{t})$ and generator $G(z; \phi_{t})$. This simply involves redefining the true distribution $(X, Z)$ by first sampling $t_\ell$ from $\{t_1, \dots, t_L\}$ and then generating $Z \sim N(0, I)$ and $X_K = Y_K + G(Z; \phi_{t_\ell})$ where $Y | Z$ follows the energy $H(G(Z; \phi_{t_\ell}) + Y ; \theta_{t_\ell})$. In this case the loss \eqref{eqn:coop_update_hat}, after replacing $\phi_t$ with $\phi_{t_\ell}$, is a stochastic approximation of the Maximum Likelihood gradient defined by the joint distribution $(t_\ell, Z, X_K)$ (see Appendix~\ref{app:hist_gen}). In practice, we implement this procedure by keeping a persistent bank of 10,000 pairs $(X, Z)$ created from past hat network updates. When updating the generator, we draw $n=128$ pairs from the bank and replace it with a newly generated batch of $n=128$ pairs from the current hat EBM. This ensures that the selection $\{t_1, \dots, t_L\}$ of past epochs remains within a close range of the current epoch $t$ with high probability. Saving the generated images $X$ at each EBM update allows us to learn from past generator weights $\phi_{t_\ell}$ without maintaining a copy of the weights. See Figure~\ref{fig:cond-hat-ebm} for an illustration and Appendix~\ref{app:algorithm} for a code sketch.

\section{Experiments}

In the subsequent empirical evaluations, we will address the following questions:

\begin{enumerate}[leftmargin=*]
\item \textbf{Refinement:} To what extent can our method refine samples from a pretrained generator model with known prior distribution?
\item \textbf{Retrofit:} Is our method capable of turning a generator model pretrained as a deterministic autoencoder into a generative model for which samples resemble realistic images?
\item \textbf{Synthesis:} Can our method learn a generator from scratch with competitive quality of synthesis on common image datasets? Can our method be scaled up to challenging datasets such as ImageNet with competitive synthesis for unconditional sampling?
\item \textbf{Out-of-Distribution:} Can the Hat EBM be used for Out-of-Distribution (OOD) detection to distinguish between samples from the training distribution and samples from dissimilar distributions?
\end{enumerate}

Figure~\ref{fig:viz_fig} visualizes some representative sampling paths for the models trained to investigate questions 1, 2, and 3 on the CIFAR-10 dataset. The appendix contains details such as pseudocode, training parameters, and model architectures. 

\subsection{Refinement}
\label{sec:exp_refinement}

In this section, we examine the problem of refining samples from a pretrained generator using a joint Hat EBM trained according to the method in Section~\ref{sec:hat_joint}. A pretrained SN-GAN with generator is used as a baseline generator $G$. We learn a Hat EBM that incorporates the generator network to refine the initial generator samples. The hat network has the exact same structure as the SN-GAN discriminator except that we remove spectral norm layers. We use fixed batch norm statistics for the generator and our energy is deterministic. The experiment is performed for the CIFAR-10 $32\times 32$ and CelebA $64\times 64$ datasets. To evaluate our method, we compare with Discriminator Driven Latent Sampling (DDLS) \cite{che2020your}, which takes the pretrained discriminator $D$ learned with $G$ and samples from the potential $U(z) = D(G(z)) + \frac{1}{2} \| z \|_2^2$. We train $D$ and $G$ using the Mimicry repository for reproducible GAN experiments \cite{lee2020mimicry} to obtain stronger baselines for $G(z)$ than those used in \cite{che2020your}.

Table~\ref{tab:refine} shows our results. The Hat EBM can learn a hat network capable of refining a image appearance more successfully than DDLS on both CIFAR-10 and CelebA. We use the joint version of the Hat EBM where both $Y$ and $Z$ are updated during sampling. Sampling is initialized from $Y_0 = 0$ and $Z_0 \sim N(0, I)$. The hat networks learns to tilt $Z$ away from its initial normal distribution to find a nearby latent vector with a more realistic appearance.

\begin{table}[h]
\caption{Improvement in FID by refining samples from a fixed SN-GAN generator.}
\label{tab:refine}
\vspace{0.25cm}
\centering
\small{
\begin{tabular}{ccc}
\toprule
Model & CIFAR-10 & CelebA \\
\midrule
SN-GAN~(baseline)~\cite{sngan} & 18.58 $\pm$ 0.08 & 6.13 $\pm$ 0.03 \\
DDLS~\cite{che2020your} & 14.59 $\pm$ 0.07 & 6.06 $\pm$ 0.01 \\
Hat EBM (\emph{Ours}) & \textbf{14.04 $\pm$ 0.11} & \textbf{5.98 $\pm$ 0.02} \\
\bottomrule
\end{tabular}
}
\end{table}

\begin{figure*}
    \centering
    \includegraphics[width=.85\textwidth,trim={0.5cm, 0cm, 0, 0},clip]{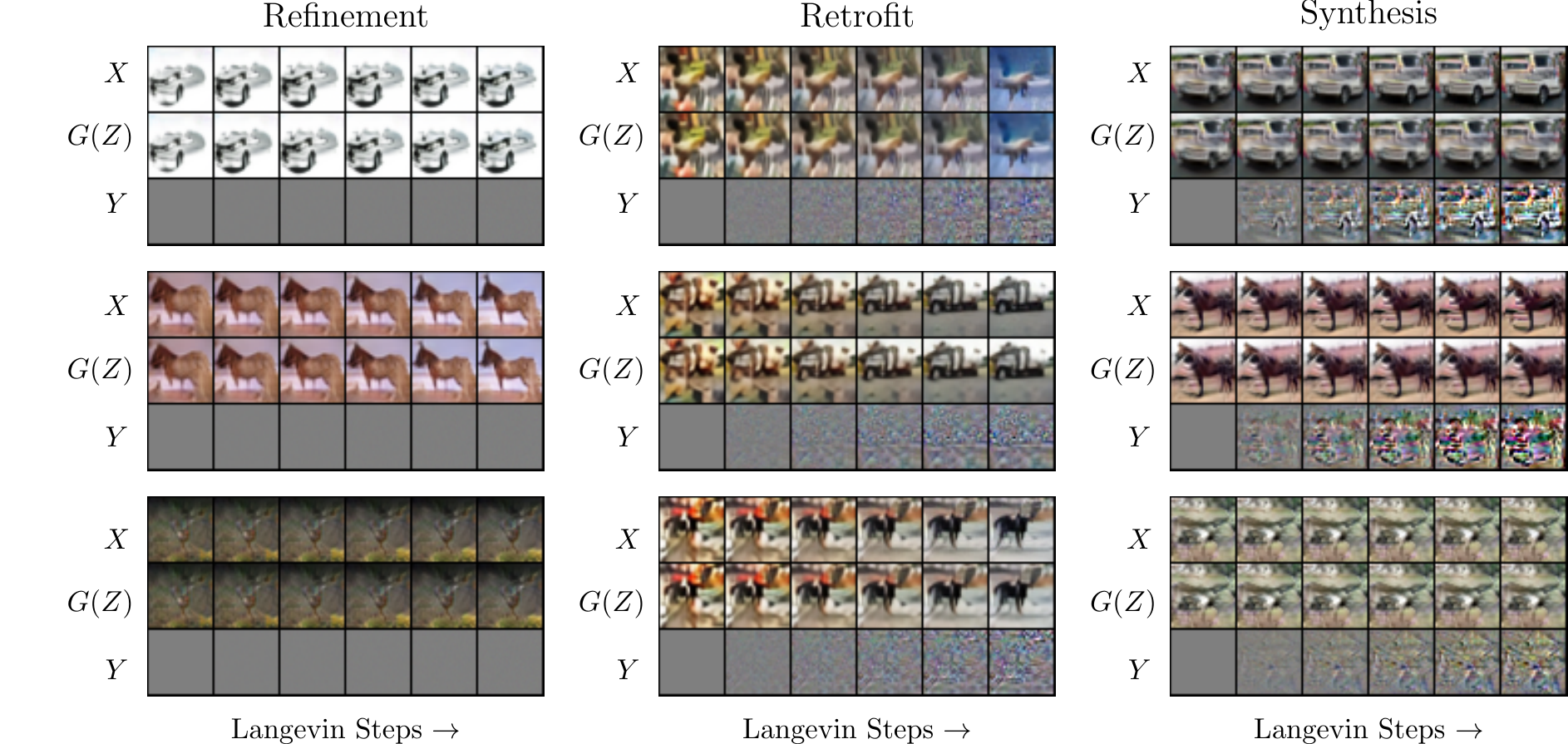}
    \caption{Visualization of shortrun Langevin paths used by each kind of Hat EBM. The same trajectories are used for both training and test-time generation. In each grouping, the three rows correspond to image $X$, generator output $G(Z)$, and residual image $Y$. All paths are initialized with residual $Y_0 = 0$ which corresponds to a grey image. For visualization purposes, all $Y$ images have been magnified by a factor of 5. The Refinement paths (\emph{left}) use the joint Hat EBM to refine the appearance of a pretrained SN-GAN generator. The residual $Y$ is barely noticeable, but still is essential for stable learning. The Retrofit paths (\emph{center}) use the joint Hat EBM to sample using a non-probabilistic autoencoder generator. The residual $Y$ is somewhat noticeable after magnification but most of the image appearance comes from $G(Z)$. The Synthesis paths (\emph{right}) use the conditional Hat EBM to sample refine a generator learned in tandem. $G(Z)$ is fixed during sampling. The residual $Y$ plays a significant role in refining the generator appearance.}
    \vspace*{-2mm}
    \label{fig:viz_fig}
\end{figure*}

We observe that the majority of the refinement is occurring in the latent space, and that the residual image $Y$ is essentially imperceptible (see Figure~\ref{fig:viz_fig}). We also notice that training quickly becomes unstable when $Y$ is removed from the Hat EBM (see Appendix~\ref{app:res_stability}). In our experience, it is essential to incorporate the residual for stable learning. A possible reason for this phenomenon is that the hat network can learn to discriminate between generator images and images not from the generator, whether they are realistic or not. If so, the hat network can assign increasingly high energy to generator samples in the absence of the residual $Y$. Even an imperceptible $Y$ is sufficient to prevent the hat network from easily distinguishing positive and negative samples so that learning is stable.

\subsection{Retrofit}

In this section, we incorporate a non-probabilistic generator network $G(z)$ into a probabilistic Hat EBM model. This essentially allows us to sample from the latent space of $G(z)$ to find latent samples whose mapping corresponds to a realistic image. Like the results in Section~\ref{sec:exp_refinement}, the residual image has small norm in the image space and most of the appearance of the sampled images comes from $G(z)$. This happens naturally without the need to coerce $Y$ to be close to 0 by including a prior term such as $p_0(y) \propto \exp\{-\frac{1}{2\sigma^2}\| y\|^2_2\}$, although including a prior term can further limit growth of $Y$.

The autoencoder generator $G(z)$ is pretrained as the second half of a standard inference network and generator network pairing. An image $X$ is fed into the inference network $I$ and converted into a latent state $Z = I(X)$, which is then decoded to obtain a reconstruction $\hat{X} = G(Z)$. The inference network and generator are learned jointly using the MSE loss $\| \hat{X} - X \|_2^2$. To keep the latent space mapping of $I(X)$ numerically stable, we project the $m$-dimensional raw output of the inference network to the sphere around the origin with radius $\sqrt{m}$ so that $\| I(X) \|_2 = \sqrt{m}$. More sophisticated methods such as perceptual and adversarial loss could have been used to train the autoencoder, but we use MSE loss to keep our implementation minimal. We observe that when $Z$ is a vector-shaped latent state, it can be difficult to learn reconstructions $\hat{X}$ with sharp appearance even for simple datasets like CIFAR-10. To obtain better reconstructions and therefore a latent space with more realistic mappings to the image space, we use image-shaped latent states $Z$. The details of our autoencoder networks can be found in Appendix~\ref{app:architectures}. When $Z$ is an image shaped latent, we treat it exactly the same as a vector latent in the learning and sampling algorithms.

We experiment with assimilating an autoencoder generator into a Hat EBM potential for the CIFAR-10 dataset. Our results are visualized in Figure~\ref{fig:viz_fig} with additional results in the Appendix~\ref{app:synthesis}. We train the Hat EBM using shortrun learning in the latent and image space by initializing $Y_0=0$ and $Z_0\sim N(0, I)$ and using $K=100$ MCMC steps of \eqref{eqn:langevin-hat-ebm-1} and \eqref{eqn:langevin-hat-ebm-2} from initialization during both training and testing evaluation to generate samples. During the Langevin dynamics, image appearance is refined mostly in the latent space. Our best model achieves a solid FID score of 26.01 $\pm$ 0.09. This demonstrates that Hat EBM can learn a probabilistic model over a non-probabilistic latent space.

\subsection{Synthesis}
\label{sec:exp_synthesis}

In this section, we use the conditional Hat EBM formulation from Section~\ref{sec:hat_tandem} to learn a hat network and generator network from scratch for self-contained synthesis. We explore synthesis for unconditional CIFAR-10 at resolution $32\times 32$, CelebA at resolution 64$\times$64, and unconditional ImageNet at resolution 128$\times$128. While recent generative models show promising results for class conditional sampling, unconditional sampling with high quality synthesis remains a significant challenge. We find especially strong results for unconditional ImageNet synthesis using the Hat EBM. This demonstrates strong potential of our synthesis method for learning with highly diverse unstructured datasets.

\begin{table}
\caption{Comparison of FID scores among representative generative models. For CIFAR-10 and Celeb-A, EBM models are above the dividing line and other generative models are below. For ImageNet, smaller scale nets are above the line and larger scale nets are below the line. (*=EBM, $\dagger$=reimplementation)}
\label{tab:fid_table}
\vspace{0.25cm}
\centering
\small{
\begin{tabular}{cc}
\begin{tabular}{ccc}
\multicolumn{2}{c}{\textbf{CIFAR-10 ($32\times 32$)}}\\
\toprule
Model  & FID \\
\midrule
Hat EBM (\emph{Ours})* & \textbf{19.30 $\pm$ 0.15} \\
Improved CD EBM \cite{du2020improved}* & 25.1 \\
VERA \cite{nomcmcforme}* & 27.5 \\
GEBM \cite{arbel2021generalized}*$\dagger$ & 30.1 \\
Cooperative EBM\cite{xie_coop}* & 33.6 \\
Flow EBM \cite{gao2020flow}* & 37.3 \\
JEM \cite{grathwohl2019modeling}* & 38.4 \\
IGEBM \cite{du2019implicit}* & 40.6 \\
\hline
DDPM \cite{ho2020denoising}  & \textbf{3.2} \\
NCSNv2\cite{song2020improved} & 10.9 \\
BigGAN \cite{biggan} & 14.7 \\
SN-GAN \cite{sngan}$\dagger$ & 18.6 \\
\bottomrule
\end{tabular}
&
\begin{tabular}{c}
\begin{tabular}{ccc}
\multicolumn{2}{c}{\textbf{CelebA ($64\times 64$)}}\\
\toprule
Model & FID \\
\midrule
Hat EBM (\emph{Ours})* & \textbf{11.57 $\pm$ 0.04} \\
Divergence Triangle \cite{han2019divergence}* & 31.9 \\
\hline
SN-GAN \cite{sngan} & \textbf{6.1} \\
NCSNv2\cite{song2020improved} & 10.2 \\
\bottomrule
\end{tabular}\\

\\
\begin{tabular}{ccc}
\multicolumn{2}{c}{\textbf{ImageNet ($128\times 128$)}}\\
\toprule
Model & FID \\
\midrule
Hat EBM (\emph{Ours})*  & \textbf{40.24 $\pm$ 0.18} \\
SN-GAN \cite{sngan}  & 65.7 \\
SS-GAN \cite{chen2019self} & 43.9 \\
InfoMax GAN \cite{lee2021infomax} & 58.9 \\
\hline
Hat EBM, scaled (\emph{Ours})* & 29.37 $\pm$ 0.15 \\
SS-GAN, scaled \cite{chen2019self} & \textbf{23.4} \\
\bottomrule
\end{tabular}
\end{tabular}
\end{tabular}
}
\vspace*{-.65cm}
\label{tab:shortrun_synthesis}
\end{table}

Our Hat EBM models use SN-GAN architectures for sizes $32\times32$, $64\times64$, and $128\times128$, where the discriminator architecture is used for the hat network. During learning, we keep the generator batch norm parameters fixed to mean 0 and variance 1. We remove all spectral norm layers from the hat network. Training parameters can be found in Appendix~\ref{app:hyperparameters}. For ImageNet models, we found that annealing the generator and hat network learning rate by a factor of 10 after 250K weight updates for each network improved the FID score significantly. See Appendix~\ref{app:synthesis} for uncurated samples from our Hat EBM models. FID results for our model and a representative selction of generative models are shown in Table~\ref{tab:fid_table}. We were unable to reproduce the GEBM FID scores reported in \cite{arbel2021generalized} and we report the score obtained from a reimplementation with the official training code. For the SN-GAN FID score, we report the stronger baseline from our reimplementation in Section~\ref{sec:exp_refinement}.

Results show strong performance of the Hat EBM compared to competing generative models across all datasets, with an especially strong performance for ImageNet. Our method significantly outperforms other EBM learning methods on CIFAR-10. The Hat EBM synthesis results are on par with the SN-GAN baseline for CIFAR-10 and CelebA, and the Hat EBM significantly outperforms SN-GAN for ImageNet. With a budget of 8 GPUs and about 2.5 days of computing, our Hat EBM achieves an ImageNet FID score of 40.0, outperforming the small-scale SS-GAN.

To our knowledge, the current state-of-the-art model for unconditional ImageNet synthesis at resolution $128\times128$ is the large-scale SS-GAN \cite{chen2019self}, which achieves an FID score of 23.4. This model was trained using a BigGAN architecture and 128-core TPUv3 pods. To scale up our Hat EBM, we doubled the number of channel dimensions for both the hat network and generator network from the original SN-GAN architecture and trained on 32-core TPU-v3 pods. Our best FID score for unconditional ImageNet 128$\times$128 was 29.2, which comes within a competitive range of state-of-the-art. We believe that further scaling in future work could enable Hat EBM to match or surpass state-of-the-art. Our results decisively demonstrate the potential of EBM learning for high-quality synthesis well beyond the scale investigated in any prior EBM work.

\subsection{Out-of-Distribution Detection}

Experiments in this section assess Hat EBM performance on Out-Of-Distribution (OOD) detection. We use the conditional Hat EBM model trained on CIFAR-10 from Section~\ref{sec:exp_synthesis} and calculate the energy $H(X; \theta)$ on unseen in-distribution images from the CIFAR-10 test set and images from OOD datasets which include CIFAR-100, CelebA, and SVHN. We follow standard OOD evaluation from works such as \cite{nalisnick2018do} which compute the AUROC metric on the energy scores of the in-distribution and OOD samples. This metric measures the ability of the Hat EBM to distinguish between in-distribution samples not seen during training and OOD samples. Following \cite{grathwohl2019modeling, xiao2021vaebm}, we expect that the energy of the OOD datasets will be higher than the energy of in-distribution test images since higher energy samples should appear with lower frequency in the learn density.

Our results are shown in Table~\ref{tab:ood_table}. The Hat EBM shows strong performance as an OOD detection method. Among methods that are fully unsupervised, our model has the top performance across all three OOD datasets. Our method approaches the results of methods that are trained with labeled data such as HDGE ~\cite{liu2021hybrid} and the fine-tuned OOD EBM~\cite{liu2020energy}, although we do not yet match these scores. Overall, there is strong evidence that the Hat EBM is naturally an effective method for OOD detection, especially when supervised label information is unavailable.

\begin{table}
\caption{Comparison of OOD scores (AUROC) among representative models. We separate scores for fully unsupervised models (above the line) and models which used supervised data (below the line).}
\label{tab:ood_table}
\vspace{0.25cm}
\centering
\small{
\begin{tabular}{cccc}
\toprule
Model & SVHN & CIFAR-100 & CelebA \\
\midrule
\textbf{Ours} & \textbf{0.92} & \textbf{0.87} & \textbf{0.94} \\
IGEBM \cite{du2019implicit} & 0.43 & 0.54 & 0.69 \\
VAEBM \cite{xiao2021vaebm} & 0.83 & 0.62 & 0.77 \\
Improved CD EBM \cite{du2020improved} & 0.91 & 0.83 & -- \\
\hline
JEM \cite{grathwohl2019modeling} & 0.67 & 0.87 & 0.77 \\
HDGE \cite{liu2021hybrid}  & 0.96 & 0.91 & 0.80 \\
OOD EBM \cite{liu2020energy} & 0.91 & 0.87 & 0.78 \\
OOD EBM (fine-tuned) \cite{liu2020energy} & \textbf{0.99} & \textbf{0.94} & \textbf{1.00} \\
\bottomrule
\end{tabular}
}
\vspace*{-.35cm}
\end{table}

\section{Conclusion}

Maximum Likelihood learning of EBMs poses a significant challenge: drawing negative samples from the current density model, which is often highly multi-modal. Prior art addresses this challenge by recruiting approximations of the EBM in the form of an ancestral sampling from a generator model, truncated Langevin chains, flow-based models, or lifting the EBM into the induced latent space of generator models. In contrast, our work proposes a method for absorbing any generator as a backbone of an EBM. The formulation assumes that observed images are the sum of unobserved latent variables pushed forward through the generator and a residual random variable which closes the gap between generator samples and image manifold. The hat network sits atop the generator and residual and both nets form the Hat EBM. The generator allows for efficient sampling but may only capture the coarse structure of the images, while the residuals can capture fine-grain details.

The Hat EBM formulation is presented in three variations: (1) joint learning of latent and residual image for adapting any fixed generator, (2) conditional learning for generators with known prior distribution, (3) self-contained learning of both EBM and generator from scratch. 
Notably, the training requires neither the log determinant of the generator Jacobian or inference of latent variables, making the learning simple and scalable.

Empirical evaluations demonstrate the various capabilities of the Hat EBM: (1) strong performance for the ImageNet synthesis at 128$\times$128 resolution with self-contained learning, (2) significant refinement of the quality of synthesis of pretrained generators on CIFAR-10 and CelebA with conditional learning, (3) retrofitting pretrained autoencoder generators with a means of sampling, (4) out-of-distribution detection with state-of-the-art performance for unsupervised models.

\section*{Acknowledgements}

This work is supported with Cloud TPUs from Google's Tensorflow Research Cloud (TFRC).

\bibliography{references}
\bibliographystyle{plain}


\section*{Checklist}

\begin{enumerate}
\item For all authors...
\begin{enumerate}
  \item Do the main claims made in the abstract and introduction accurately reflect the paper's contributions and scope?
    \answerYes{}
  \item Did you describe the limitations of your work? 
    \answerYes{} \textcolor{blue}{See Appendix~\ref{app:limitations}.}
  \item Did you discuss any potential negative societal impacts of your work?
    \answerYes{} \textcolor{blue}{See Appendix~\ref{app:ethics}.}
  \item Have you read the ethics review guidelines and ensured that your paper conforms to them?
    \answerYes{}
\end{enumerate}

\item If you are including theoretical results...
\begin{enumerate}
  \item Did you state the full set of assumptions of all theoretical results?
    \answerYes{} \textcolor{blue}{See Appendix~\ref{app:validity}}.
        \item Did you include complete proofs of all theoretical results?
    \answerNA{}
\end{enumerate}

\item If you ran experiments...
\begin{enumerate}
  \item Did you include the code, data, and instructions needed to reproduce the main experimental results (either in the supplemental material or as a URL)?
    \answerYes{} \textcolor{blue}{Code and pretrained models are available at \url{https://github.com/point0bar1/hat-ebm}.}
  \item Did you specify all the training details (e.g., data splits, hyperparameters, how they were chosen)?
    \answerYes{} \textcolor{blue}{See configs in code files and Appendix~\ref{app:hyperparameters}.}
        \item Did you report error bars (e.g., with respect to the random seed after running experiments multiple times)?
    \answerYes{} \textcolor{blue}{Primary numerical results are FID scores. Each experiment was run 3 times and estimated standard deviation was reported.}
        \item Did you include the total amount of compute and the type of resources used (e.g., type of GPUs, internal cluster, or cloud provider)?
    \answerYes{} \textcolor{blue}{See Appendix~\ref{app:resources}.}
\end{enumerate}

\item If you are using existing assets (e.g., code, data, models) or curating/releasing new assets...
\begin{enumerate}
  \item If your work uses existing assets, did you cite the creators?
    \answerYes{} \textcolor{blue}{Our code is annotated to acknowledge our use of existing assets. In particular, we used network structures based on existing popular models. All other code was created by the authors.}
  \item Did you mention the license of the assets?
    \answerNA{}
  \item Did you include any new assets either in the supplemental material or as a URL?
    \answerYes{} \textcolor{blue}{Code and pretrained models are available at \url{https://github.com/point0bar1/hat-ebm}.}
  \item Did you discuss whether and how consent was obtained from people whose data you're using/curating?
    \answerNA{}
  \item Did you discuss whether the data you are using/curating contains personally identifiable information or offensive content?
    \answerNA{}
\end{enumerate}

\item If you used crowdsourcing or conducted research with human subjects...
\begin{enumerate}
  \item Did you include the full text of instructions given to participants and screenshots, if applicable?
    \answerNA{}
  \item Did you describe any potential participant risks, with links to Institutional Review Board (IRB) approvals, if applicable?
    \answerNA{}
  \item Did you include the estimated hourly wage paid to participants and the total amount spent on participant compensation?
    \answerNA{}
\end{enumerate}
\end{enumerate}

\appendix

\section{Limitations}\label{app:limitations}

A major limitation of our method is the computational cost of MCMC sampling. Updating the EBM weights requires about 50 to 100 MCMC steps, and each step requires a backward pass to compute the Langevin gradient. Despite this costly step, training our model requires substantially less resources than GANs that achieve similar quality results (see Appendix~\ref{app:resources}). The joint version of the Hat EBM further increases computational requirements compared to standard EBM because of the need for dual Langevin updates and the need to backprop through the generator. The conditional Hat EBM only requires a backprop through the hat network and forward pass through the generator, which has similar runtime to standard EBMs.

Another limitation of our work is that we still rely on noise-initialized shortrun sampling for our retrofit experiments, which requires that the initialization distribution in the latent space is a reasonable starting point for obtaining good latent space samples. This is done in our work by enforcing that latent vectors corresponding to images lie on a sphere of radius $\sqrt{m}$, so that sampling from a $m$-dimensional Gaussian is roughly aligned with the target states. In future work, we hope to develop a better latent space initialization method, perhaps by adapting cooperative learning to give latent space initialization rather than image space initialization.

\section{Potential Negative Impacts}\label{app:ethics}

Like many works in generative modeling, our work has the potential to contribute to the development of harmful images, which could take the form of images that spread misinformation, explicit content, or images that perpetuate negatives biases and stereotypes. 

\section{Computational Requirements}\label{app:resources}

Our computational resources were primarily 5 TPUv2-8 and 5 TPUv3-8 devices. For our large-scale ImageNet experiments, we used a TPUv3-32 device. Our large scale ImageNet experiment can be run on a TPUv3-32 in approximately 60 hours, coming to a total of $32 \times 60 = 1920$ TPU hours. We were also able to run our largest scale experiment on a TPUv3-8 in approximately 130 hours, coming to a total of $8 \times 130 = 1040$ TPU hours for the same experiment. Since we only get about a 2$\times$ speed-up when parallelizing from 8 to 32 cores, the overall compute is lower for the TPUv3-8, although the actual runtime is longer. This runtime compares favorably with the state-of-the-art Self-Supervised GAN (SSGAN) \cite{chen2019self} for unconditional ImageNet 128$\times$128 synthesis, which reports a runtime of about 1.5 days using a TPUv3-128 and a cost of $36 \times 128 = 4608$ TPU hours. Despite the costly step of MCMC sampling, our relatively lightweight networks and parallelization of Langevin sampling make EBM learning feasible.

To estimate our total amount of compute, we use the following calculation. We performed experiments for this work over the course of 3 months. We estimate that on average, we had approximately 4 TPU-8 machines (either v2 or v3) running at any given time. This work done by the TPU-8 machines comes to about $8 \times 4 \times 3 \times 730 = 70080$ TPU hours. We additionally performed about 15 experiments using the TPUv3-32, which comes to about $15 \times 1920 = 28800$ TPU hours. In total, we used approximately 100K TPU hours across all experiments.

\section{Validity of Learning Procedures and Assumptions}\label{app:validity}

Our primary theoretical claim is the validity of the joint Hat EBM and conditional Hat EBM learning procedures. The validity of the learning procedures relies on two key assumptions: correct model specification for the observed data, and the convergence of MCMC samples during the training to ensure that negative samples represent samples from the current model. 

Our assumptions about the distribution of the observed data $X$ are as follows. For the joint Hat EBM, we assume that $X \sim Y + G(Z)$ for a joint distribution $(Y, Z)$ that can be parameterized as $U(y, z; \theta) = H(Y + G(Z); \theta)$. Since $Y$ is unconstrained, it seems reasonable that for a suitably flexible function $H$, one could always learn a residual that corrects the appearance of any generator, even a poor generator. Thus the model is well-specified since $Y+ G(Z)$ should be able to represent an arbitrary appearance regardless of the generator $G$ if $H$ has sufficient capacity. For the conditional Hat EBM, the observed data $X$ are assumed to follow $X = Y + G(Z)$ where $Z\sim N(0,I)$ and $U(Y|Z; \theta) = H(Y + G(Z); \theta)$. Again, as long as $H$ has sufficient capacity, it should be able to learn residuals $Y$ that correct the appearance of any generator.

The second assumption requires that MCMC samples converge to their steady-state to update the Hat Network. It is known from previous work that the predominant outcome of most MCMC sampling during EBM training is very far from convergence~\cite{nijkamp2019anatomy}. Our work operates exclusively in the non-convergent regime, and we acknowledge that the second assumption is violated. Prior work has identified that realistic synthesis is much more easily achieved in the non-convergent regime, and violating the convergence assumption is standard practice in EBM learning with the goal of synthesis.

\section{Architectures}\label{app:architectures}

All architectures except for the retrofit generator use a standard SN-GAN discriminator or generator architecture. The hat network uses a discriminator structure with spectral norm layers removed. Batch norm is set to test mode for all generators. The scaled ImageNet experiments simply double the scaling factor for channel dimensions of the EBM and Generator from 1024 to 2048 for the 128$\times$128 SN-GAN architecture. The generator used in the CIFAR-10 retrofit experiment has a image-style latent space with dimension [16, 16, 1]. This network has the same architecture has the SN-GAN generator except that the it removes the fully connected layer at the base of the generator and replaces it with a convolutional layer, and the first two upsampling residual blocks in the original SN-GAN are converted to residual blocks that do not upsample.

\section{Historical Generator Update}\label{app:hist_gen}

A key aspect of our self-contained learning procedure using the Conditional Hat EBM is the use of historical EBM samples to update the generator, rather than samples from the current EBM. This dramatically improves the quality of synthesized samples. We hypothesize that using only current EBM samples to update the generator fails because the short MCMC trajectories cannot significantly change the appearance of the initial generator samples. This means that a lack of diversity in the initial generator samples will also lead to a lack of diversity of samples used to update the EBM. Lack of sample diversity can easily cause EBM instability because the EBM will rapidly change its landscape to try to cover modes that are not contained in the negative samples, causing it to forget previous modes. Once it has forgotten a previous mode, it will once again experience a rapid update to recover the forgotten mode, at the expense of forgetting another mode. Stable EBM learning requires that the initial states for MCMC sampling have reasonable diversity so that learning in many modes can take place simultaneously. We find the same problem occurs for the original cooperative learning formulation \cite{xie_coop}.

\begin{figure}[ht]
    \centering
    \begin{tabular}{cccc}
    \includegraphics[width=.19\textwidth]{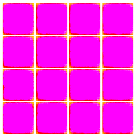} &
    \includegraphics[width=.19\textwidth]{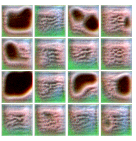} &
    \includegraphics[width=.19\textwidth]{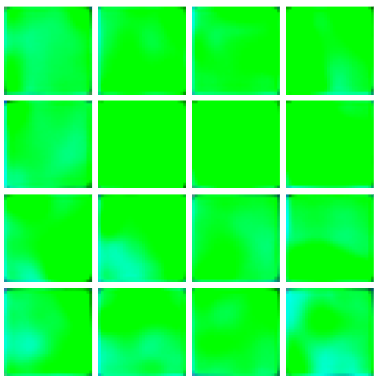} &
    \includegraphics[width=.19\textwidth]{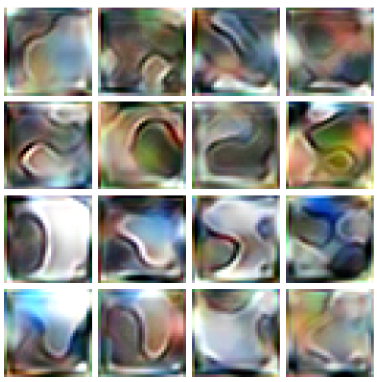} \\
    Coop. (no batch norm) & Coop. (batch norm) & Hat EBM (current) & Hat EBM (historical)
    \end{tabular}
    \caption{Samples after 500 weight updates for different EBM learning methods. Cooperative learning fails without batch norm because the tether between the EBM and generator leads to lack of diversity and instability. Including batch norm in cooperative learning helps add some diversity, but samples can still remain very visually similar for many updates. The Hat EBM has similar problem as cooperative learning when current EBM samples are used to update the generator. This problem is greatly alleviated by instead using historical samples to update the generator, because several past updates of the EBM have much higher diversity than a single snapshot.}
    \label{fig:historical}
\end{figure}

We visualize the importance of our historical update in Figure~\ref{fig:historical}. This figure compares cooperative learning \cite{xie_coop} with and without batch normalization to Hat EBM synthesis for CIFAR-10. One Hat EBM experiment uses only the current EBM to update the generator, while the other uses the historical approach we outline in the text. Neither of our Hat EBM experiments use batch norm. The Hat EBM using historical updates is by far the most successful synthesis method early in training, and we find that this advantage is maintained throughout learning. We also find that batch norm is an essential part of the original cooperative learning method because it prevents the generator from collapsing early in training.

This historical update can be viewed as performing maximum likelihood where the data distribution $(X, Z)$ is defined as the marginal of the joint distribution of $(t_\ell, X, Z)$ where $X = Y_K + G(Z; \phi_{t_\ell})$ and $Y_K | Z$ is sampled from
\small{
\[
p (Y | Z, t_\ell) \propto - \exp\{ H(Y + G(Z; \phi_{t_\ell}) ; \theta_{t_\ell}) \}.
\]
}
This holds because
\small{
\begin{align*}
    E_{(t_\ell, Z, X)} [-\log p_G (X, Z ; \phi)] &= E_{(t_\ell, Z, X)} \left[\frac{1}{2 \tau^2} \| G(Z ; \phi) - X \|_2^2 \right] + C \\
    &\approx \frac{1}{n}\sum_{i=1}^n \frac{1}{2 \tau^2} \| G(Z_i ; \phi) - (Y_{K,i} + G(Z_i ; \phi_{t_{\ell_i}})) \|_2^2 + C
\end{align*}
}
for $n$ samples $\{(t_{\ell_i}, Z_i, X_i) \}_{i=1}^n$.

\section{Algorithm for Tandem Training of Conditional Hat EBM}\label{app:algorithm}

\begin{algorithm}
\caption{Training a Conditional Hat EBM for Image Synthesis}\label{alg:coop_persistent}
\scriptsize{
\begin{algorithmic}
\title{d}
\REQUIRE{Natural images $\{x^+_m\}_{m=1}^{M}$, EBM $U(x;\theta)$, generator $G(z;\phi)$ Langevin noise $\varepsilon$, number of shortrun steps $K$, EBM optimizer $h_U$, generator optimizer $h_G$, random initial weights $\theta_0$ and $\phi_0$, number of training iterations $T$, bank size $N$.}
\ENSURE{Learned weights $\theta_T$ for EBM and $\phi_T$ for generator.}
\STATE{Initialize bank of random latent states $\{Z_i\}_{i=1}^N$ i.i.d. from the Gaussian $N(0, I)$.}
\STATE{Initialize image bank $\{X^-_i\}_{i=1}^N$ from generator using $X_i^- = g(Z_i; \phi_0)$}
\FOR{$1 \leq t \leq T$}

\medskip

\STATE{\textbf{Steps to Update EBM}}
\STATE{Select batch $\{\tilde{X}^+_b \}_{b=1}^B$ from data samples $\{x^+_m\}_{m=1}^{M}$.}
\STATE{Draw latent samples $\{\tilde{Z}_b\}_{b=1}^B$ i.i.d. from the Gaussian $N(0, I)$.}
\STATE{Initialize residual images $\{\tilde{Y}^-_{b, 0}\}_{b=1}^B$ from the image with all pixels set to 0.}
\STATE{Update residual images $\{\tilde{Y}^-_{b,0}\}_{b=1}^B$ with $K$ Langevin steps of Equation 6 to obtain $\{\tilde{Y}^-_{b,K}\}_{b=1}^B$. Keep $\tilde{Z}_b$ fixed.}
\STATE{Sum generated image and residual image using $\tilde{X}_b^- = G(\tilde{Z}_b, \phi_{t-1}) + \tilde{Y}^-_{b,K}$ to obtain negative samples $\{\tilde{X}^-_{b}\}_{b=1}^B$.}
\STATE{Get learning gradient $\Delta^{(t)}_U$ using Equation 8 with samples $\{\tilde{X}^+_{b}\}_{b=1}^B$ and $\{\tilde{X}^-_{b}\}_{b=1}^B$.}
\STATE{Update $\theta_t$ using gradient $\Delta^{(t)}_U$ and optimizer $h_U$.}

\medskip

\STATE{\textbf{Steps to Update Generator}}
\STATE{Randomly choose unique indices $\{i_1, \dots, i_B\} \subset \{1, \dots, N\}$.}
\STATE{Get paired batches $\{Z_{i_b}\}_{b=1}^B$ and $\{X^-_{i_b}\}_{b=1}^B$ from $\{Z_i\}_{i=1}^N$ and $\{X^-_i\}_{i=1}^N$.}
\STATE{Get learning gradient $\Delta^{(t)}_G$ using Equation 11 with samples $\{Z_{i_b}\}_{b=1}^B$ and $\{X^-_{i_b}\}_{b=1}^B$.}
\STATE{Update $\phi_t$ using gradient $\Delta^{(t)}_G$ and optimizer $h_G$.}
\STATE{Overwrite old states $\{Z_{i_b}\}_{b=1}^B$ and $\{X^-_{i_b}\}_{b=1}^B$ in bank with update $Z_{i_b} \leftarrow \tilde{Z}_b$ and $X^-_{i_b} \leftarrow \tilde{X}_b^-$.}
\ENDFOR
\end{algorithmic}
}
\end{algorithm}

\section{Importance of Residual Image for Stability}
\label{app:res_stability}

\begin{figure}[ht]
    \centering
    \includegraphics[width=.475\textwidth]{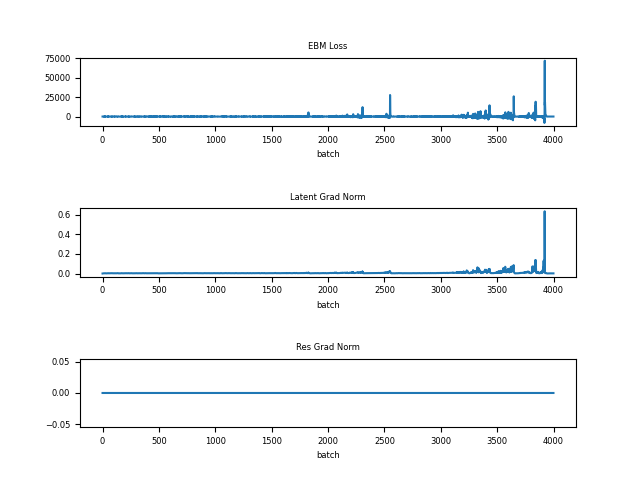} \quad \includegraphics[width=.475\textwidth]{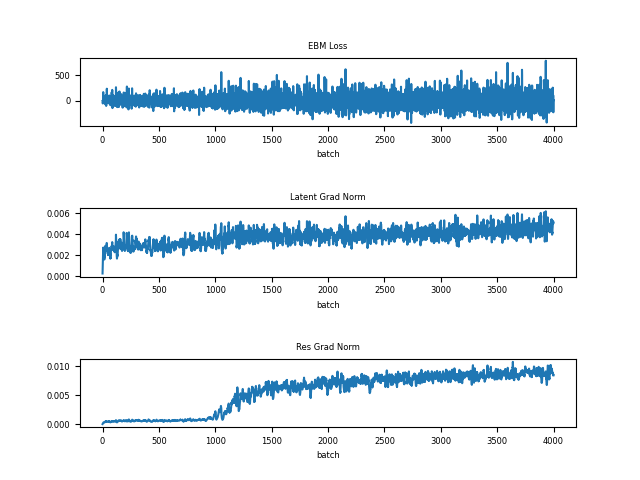}
    \caption{Unstable learning using the energy \eqref{eqn:bad_hat_ebm} \emph{(left)} and stable learning using the joint Hat EBM \emph{(right)}. Both settings replicate the refinement experiment using CIFAR-10 with a pretrained SN-GAN generator. Even though the appearance of the residual image $Y$ is nearly invisible, including the residual is essential both from a theoretical perspective and for practical stability.}
    \label{fig:stability}
\end{figure}

Throughout our experiments with different versions of the Hat EBM, we find the inclusion of the residual image $Y$ essential for stability. In particular, one could consider an alternate version of the Hat EBM where 
\begin{equation}
    U(z; \theta) = H(G(z); \theta)\label{eqn:bad_hat_ebm}
\end{equation} 
without a residual state. As long as the training data is of the form $X^+ = G(Z^+)$ for a latent state $Z^+$, one could learn the hat network using the same procedure as the Hat EBM without the residual $Y$. In practice, it is usually not possible to exactly invert the generator. In other words, real images $X^+$ never lie exactly on the generator output manifold, although they might be close by. Nonetheless, one could bend the rules and use $X^+$ to train the potential \eqref{eqn:bad_hat_ebm} with the justification that there is some $Z^+$ such that $X^+ \approx G(Z^+)$. In practice, this leads to instability as shown in Figure~\ref{fig:stability}. Even when it is nearly invisible, the residual state $Y$ is still needed for the hat network to balance the energy of positive and negative samples and achieve stable learning. 

\section{Discussion of EBM Synthesis Methods}
\label{app:ebm_discussion}

This appendix provides further discussion of related EBM methods, including methods presented in Table 2 of the main paper, and draws relevant comparisons between the Hat EBM and other EBMs. 

One branch of EBM works uses MCMC-based Maximum Likelihood with persistent initialization of MCMC states. Persistent initialization uses samples of prior short run EBM trajectories to initialize the current sampling trajectory. This approach is introduced by Persistent Contrastive Divergence (PCD) \cite{tieleman2008training}. The IGEBM \cite{du2019implicit} is trained using a bank with 10,000 images to hold persistent states. States are rejuvenated from a Gaussian or uniform noise image with of between 0.5\% and 5\% probability before being returned to the image bank. The Improved CD EBM \cite{du2020improved} builds on these results by including an approximate KL divergence term in EBM learning to minimize the difference between the data distribution and the sampled distribution, and by rejuvenating MCMC trajectories using data augmentation instead of resetting states with noise. The Joint Energy Model (JEM) \cite{grathwohl2019modeling} trains an unconditional EBM and a classifier model simultaneously with the same network using persistent initialization with noise rejuvenation. The use of persistent states in our work differs from prior work because we use persistent states to update only the generator while the EBM is updated by states generated from scratch in the current iteration. This is done to increase the diversity of samples used to update the generator, which is essential for enabling the generator to create distinct appearances for different $Z$ early in training (see Appendix~\ref{app:hist_gen}).

Another branch of EBM works trains a generator network in tandem with the energy network. Most works use the standard EBM update or a close variant to train the energy network, as we do. In some works, the generators produce the final samples and no MCMC is used, while other works use the generator to initialize samples and then refine the samples with MCMC driven by the energy network. Our work adopts the second strategy. To our knowledge, the first work that explores the idea jointly training an energy network and generator network is by Kim \& Bengio \cite{kim2016deep}. This work suggests using the generator samples directly as negative samples without use of MCMC, and updating the generator network to decrease the energy of the generator samples. The EGAN~\cite{dai2017calibrating} builds on \cite{kim2016deep} by introducing a entropy maximization term which is needed for a valid Maximum Likelihood objective and which prevents generator collapse. The entropy term is estimated by neighborhood methods and variational methods. MEG~\cite{Kumar2019MaximumEG} and VERA \cite{nomcmcforme} build on \cite{dai2017calibrating} by introducing more sophisticated methods of entropy maximization. The GEBM~\cite{arbel2021generalized} uses an approach similar to \cite{dai2017calibrating}, with the major differences being use of a generalized log likelihood objective that bridges the gap between the support of the generator output and the full image space distribution of the data, and a novel approximate KL bound for learning the generator. Like the Hat EBM, none of these methods require the log determinant of the generator Jacobian or inference of latent states for data. Unlike the Hat EBM, the probability models from these methods lie in the latent space (or the restricted image space given by the generator outputs) instead of the full image space. The methods are also incompatible with non-probabilistic generators, unlike Hat EBM. None of the works above use MCMC during training, although some use MCMC during synthesis \cite{nomcmcforme, arbel2021generalized}. Cooperative learning \cite{xie_coop} uses Maximum Likelihood learning described in Section 3.4. This requires MCMC sampling for both image and latent states. The conditional Hat EBM for synthesis requires sampling for image states but not latent states.

A third branch of EBM methods initialize MCMC sampling from a noise distribution and use a fixed length MCMC trajectory to generate states without a generator network. This branch differs from persistent methods because no persistent bank is used and negative samples to update the EBM are created from scratch each time the EBM weights are updated. It differs from generator methods because realistic synthesis is achieved through pure MCMC without initial realistic states from the generator. The Multigrid EBM \cite{multigrid} has a MCMC-based training method where images are synthesized and sampled at multiple resolutions. Multiple EBMs are learned in parallel at different resolutions, and generated images from low resolution EBMs are passed to high resolution EBMs to initialize MCMC sampling. Generation can be performed by trivial sampling (uniform, Gaussian Mixture, KDE, etc.) at a single-pixel resolution and passing the generated MCMC states along from the single-pixel EBM to the full-size EBM. The short run initialization method \cite{nijkamp2019learning} starts sampling from a uniform image distribution and runs 50 to 100 MCMC steps to generate images during each EBM update, bypassing the need for persistent banks. Our retrofit Hat EBM training is a variation of the short run method where both the $Y$ and $Z$ are initialized from uniform noise. Since the generator is non-probabilistic, the short run trajectories of $Z$ must move from uniform latent samples that represent noisy images to tuned latent samples whose generated images match the data appearance.

\section{Hyperparameters}\label{app:hyperparameters}

\bigskip
\bigskip

\begin{center}
\begin{tabular}{cccc}
\multicolumn{4}{c}{\textbf{Synthesis Training}}\\
\toprule
Dataset  & Celeb-A & CIFAR-10 & ImageNet \\
\midrule
{Training Steps} &  {50000} & {75000} & {300000} \\
{Batch Size} & {128} & {128} & {128} \\
{Data Epsilon} & {1e-3} & {1e-3} & {1e-3} \\
{EBM LR} & {1e-4} & {1e-4} & {1e-4} \\
EBM Optimizer & Adam & Adam & Adam \\
EBM Gradient Clip & None & None & 50 \\
{Langevin Epsilon} & {5e-4} & {5e-4} & {5e-4} \\
{MCMC Steps} & {50} & {50} & {50} \\
{MCMC Temperature} & {1e-8} & {1e-3} & {1e-8} \\
{Persistent Bank Size} & {10000} & {10000} & {10000} \\
{Generator LR} & {1e-4} & {1e-4} & {5e-5} \\
Generator Optimizer & Adam & Adam & Adam \\
\bottomrule
\end{tabular}

\bigskip
\bigskip

\begin{tabular}{ccc}

\begin{tabular}{cc}
\multicolumn{2}{c}{\textbf{Retrofit Training}}\\
\toprule
Dataset  & CIFAR-10 \\
\midrule
{Training Steps} &  {30000}  \\
{Batch Size} & {128} \\
{Data Epsilon} & {1e-3} \\
{EBM LR} &  {1e-4} \\
EBM Optimizer & Adam \\
{Image Space Epsilon} & {5e-4}  \\
{Latent Epsilon} & {1e-3}  \\
{MCMC Steps} & {100} \\
{MCMC Temperature} & {1e-3} \\
{Prior $\sigma$} & {0.1} \\
\bottomrule
\end{tabular}

& &

\begin{tabular}{cccc}
\multicolumn{3}{c}{\textbf{Refinement Training}}\\
\toprule
Dataset  & Celeb-A & CIFAR-10 \\
\midrule
{Training Steps} &  {20000} & {20000} \\
{Batch Size} & {128} & {128} \\
{Data Epsilon} & {1e-3} & {1e-3} \\
{EBM LR} & {1e-5} & {1e-5} \\
EBM Optimizer & Adam & Adam \\
{Image Space Epsilon} & {1e-4} & {1e-4} \\
{Latent Epsilon} & {5e-3} & {5e-3} \\
{MCMC Steps} & {100} & {250} \\
{MCMC Temperature} & {1e-6} & {1e-3} \\
{Prior $\sigma$} & {None} & {0.25} \\
\bottomrule
\end{tabular}

\end{tabular}
\end{center}

\newpage

\section{Visualization of Synthesis Results}\label{app:synthesis}

\bigskip
\bigskip
\bigskip

\begin{figure}[ht]
    \centering
    \includegraphics[width=.45\textwidth]{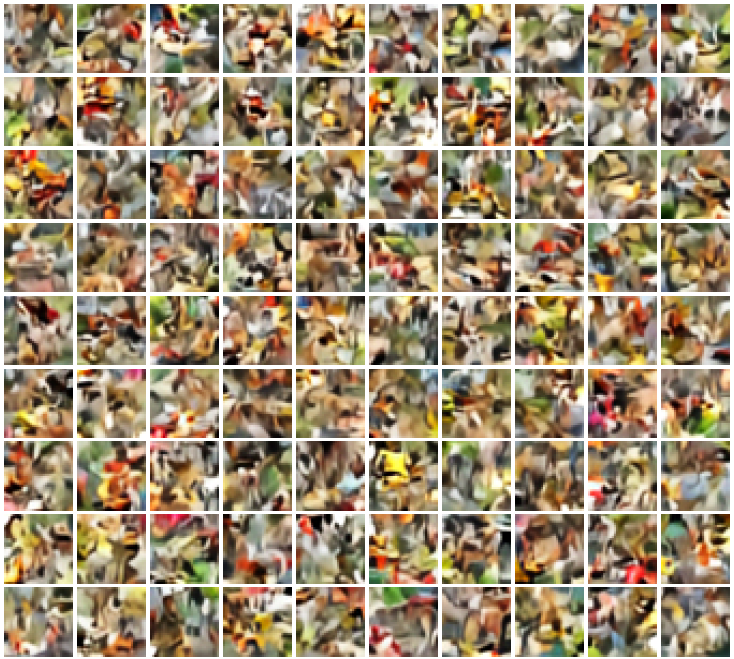} \quad \includegraphics[width=.45\textwidth]{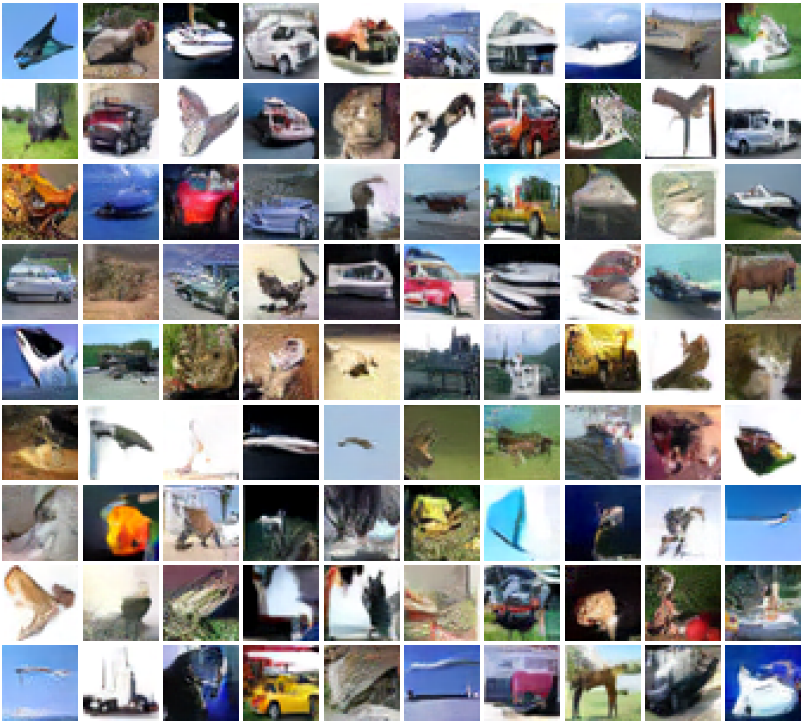}
    \caption{Initial image states \emph{(left)} and sampled image states \emph{(right)} for retrofit Hat EBM that uses a pretrained generator from a deterministic autoencoder. The training dataset is CIFAR-10.}
    \label{fig:retrofit_cifar10}
\end{figure}

\bigskip
\bigskip
\bigskip

\begin{figure}[ht]
    \centering
    \includegraphics[width=.55\textwidth]{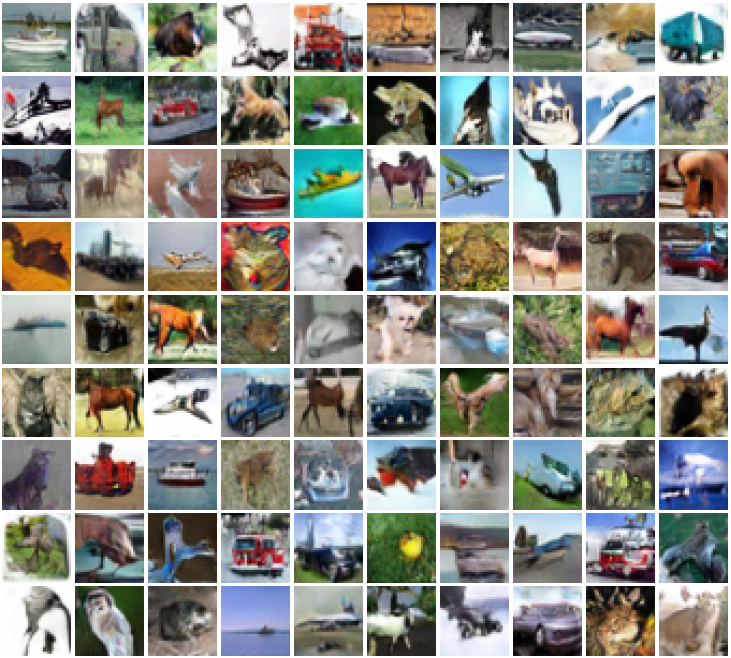}
    \caption{Uncurated Hat EBM samples for unconditional CIFAR-10 at resolution $32\times 32$.}
    \label{fig:shortrun_cifar10}
\end{figure}

\newpage

\bigskip
\bigskip

\begin{figure}[ht]
    \centering
    \includegraphics[width=.825\textwidth,trim={2.75cm, 5.05cm, 2.5cm, 4.925cm},clip]{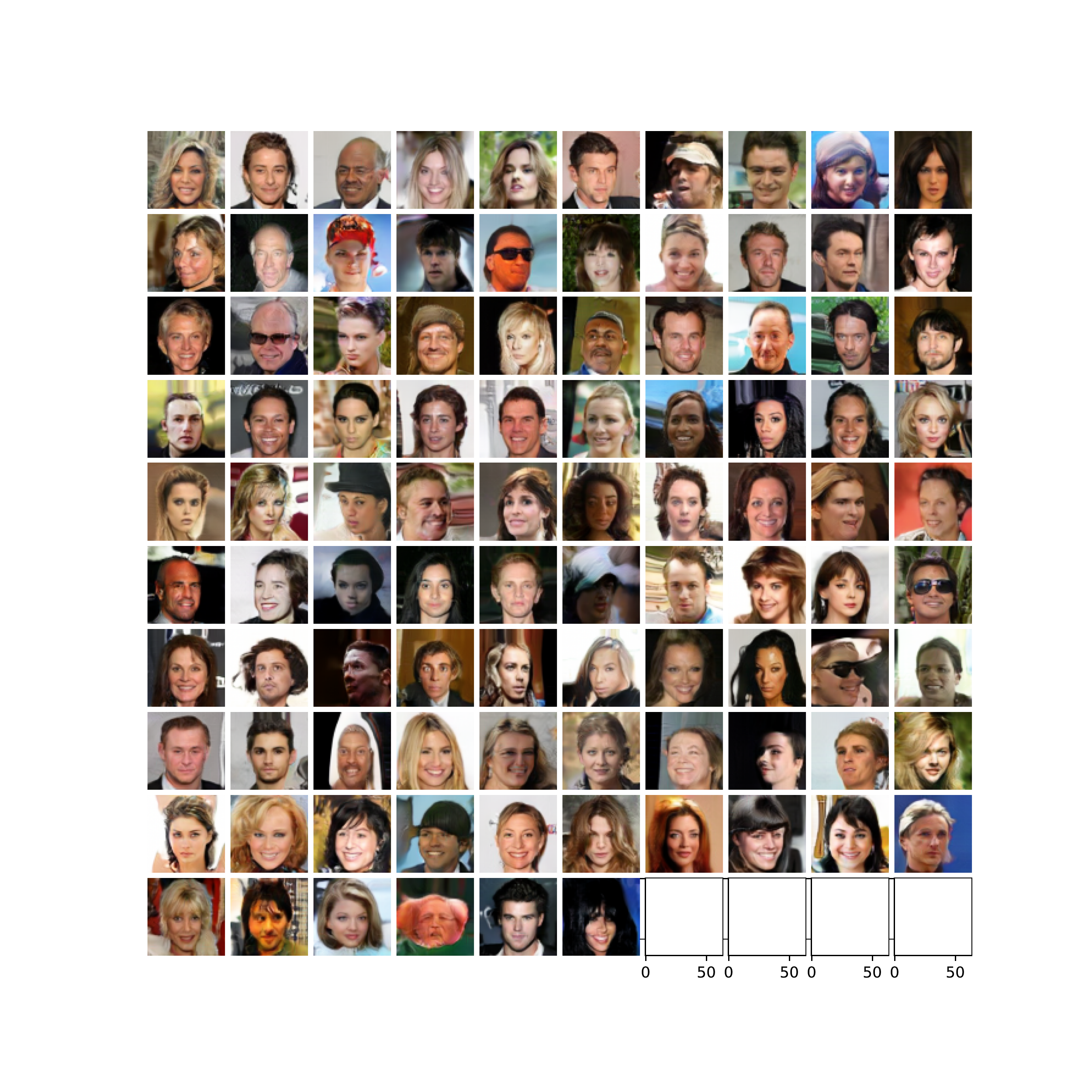}
    \caption{Uncurated Hat EBM samples for Celeb-A at resolution $64\times 64$.}
    \label{fig:shortrun_celeb-a}
\end{figure}

\bigskip
\bigskip

\begin{figure}[ht]
    \centering
    \includegraphics[width=.825\textwidth,trim={2.75cm, 5.05cm, 2.5cm, 4.925cm},clip]{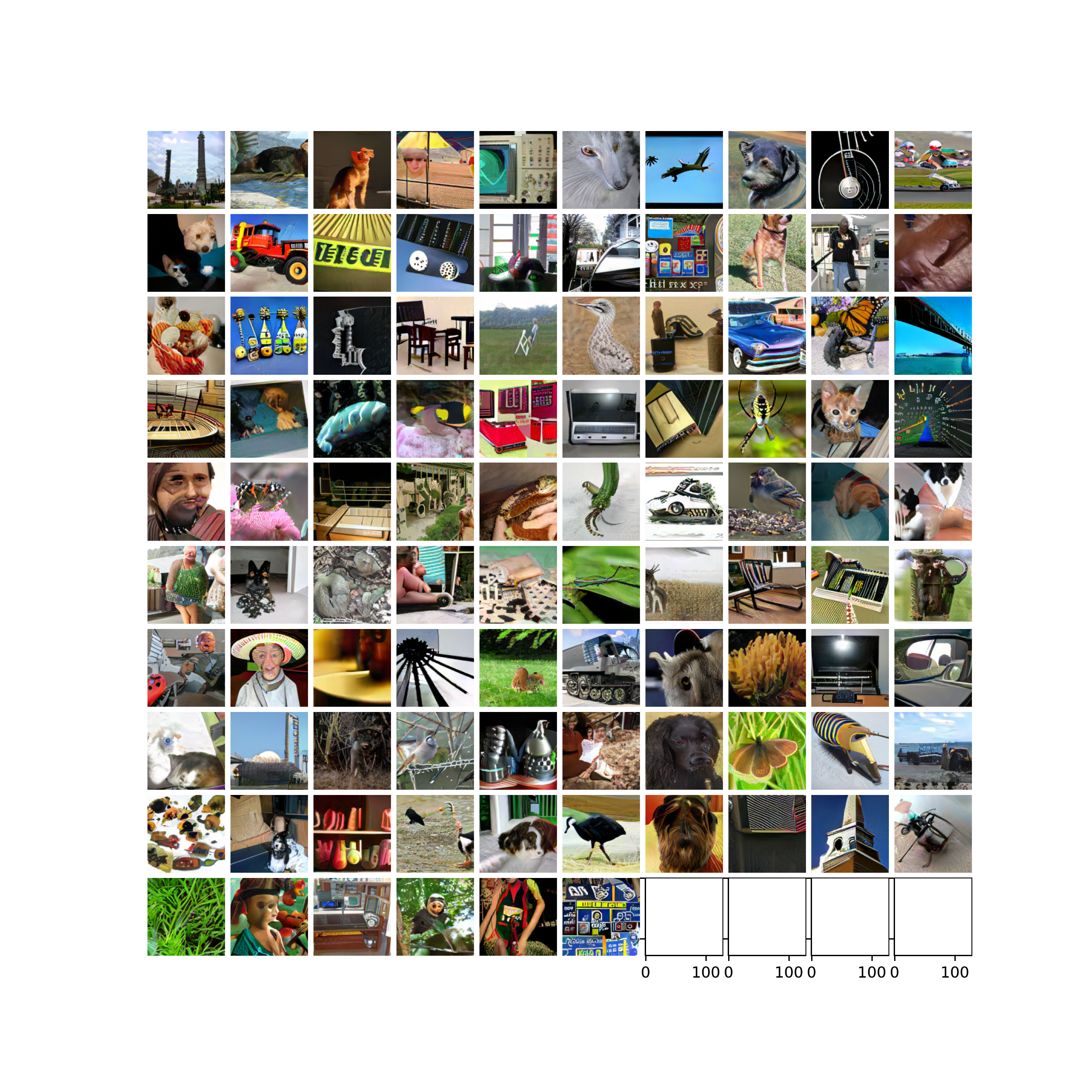}
        \caption{Uncurated Hat EBM samples for unconditional ImageNet at resolution $128\times128$.}
    \label{fig:shortrun_imagenet}
\end{figure}

\end{document}